\documentclass[letterpaper]{article}
\usepackage{flairs}%aaai
\usepackage{times}
\usepackage{helvet}
\usepackage{courier}

\usepackage{url}
\usepackage{graphicx}
\usepackage{hyperref}
\usepackage[hang]{footmisc}
\begin{document}
% This file is an adoption of the style file for AAAI Press 
% proceedings, working notes, and technical reports.  This file is made 
% with minimal changes by explicit permission from AAAI.
\title{Exploring the Structure of AI-Induced Language Change in Scientific English}
\author{}

\author{Riley Galpin\footnotemark[1], Bryce Anderson, Tom S.\ Juzek\thanks{\noindent Corresponding authors. Author contributions:\ Conceptualisation and methodology (RG, TSJ), data collection and analysis (RG, BA, TSJ), coding and visualisation (RG, TSJ), write up (RG, BA, TSJ).}\\
Florida State University\\
%Address Line 1.1\\
%Address Line 1.2\\
%\And Author 3\\
%Affiliation 3\\
%Address Line 3.1\\
%Address Line 3.2\\
}
\maketitle

\begin{abstract}
Scientific English has undergone rapid and unprecedented changes in recent years, with words such as ``delve,'' ``intricate,'' and ``crucial'' showing significant spikes in frequency since around 2022. These changes are widely attributed to the growing influence of Large Language Models like ChatGPT in the discourse surrounding bias and misalignment. However, apart from changes in frequency, the exact structure of these linguistic shifts has remained unclear. The present study addresses this and investigates whether these changes involve the replacement of synonyms by suddenly `spiking words,' for example, ``crucial'' replacing ``essential'' and ``key,'' or whether they reflect broader semantic and pragmatic qualifications. To further investigate structural changes, we include part of speech tagging in our analysis to quantify linguistic shifts over grammatical categories and differentiate between word forms, like ``potential'' as a noun vs.\ as an adjective. We systematically analyze synonym groups for widely discussed `spiking words' based on frequency trends in scientific abstracts from PubMed. We find that entire semantic clusters often shift together, with most or all words in a group increasing in usage. This pattern suggests that changes induced by Large Language Models are primarily semantic and pragmatic rather than purely lexical. Notably, the adjective ``important'' shows a significant decline, which prompted us to systematically analyze decreasing lexical items. Our analysis of ``collapsing'' words reveals a more complex picture, which is consistent with organic language change and contrasts with the patterns of the abrupt spikes. These insights into the structure of language change contribute to our understanding of how language technology continues to shape human language.
\end{abstract}

\section{1 Introduction}
\label{sec:intro}

Artificial Intelligence (AI) is a fast-growing field that has resulted in impacts on many aspects of everyday life that have yet to be comprehensively detailed \cite{makridakis2017forthcoming,mahajan2023artificial,pachegowda2023global}, including its impact on human language. Specifically, in the past few years, Scientific English has undergone considerable, rapid language change. In principle, language change is a well-documented phenomenon \cite{aitchison2005language_change}, driven by a range of factors, including (see the following works for representative discussions):\ social factors \cite{labov1973sociolinguistic,milroy1985linguistic}, language contact, \cite{thomason2023language,winford2003introduction}, cognitive constraints \cite{bybee2006usage,hopper1984discourse}, grammaticalization processes \cite{heine2017grammaticalization,haspelmath2000grammaticalization}, and technological developments \cite{crystal2008txtng,baron2010always}. 

Scientific English, too, is subject to these mechanisms of change and has been extensively studied as a domain of diachronic variation \cite{degaetano2018using,bizzoni2020linguistic,menzel2022medical,krielke2024cross}. However, the abruptness and scale of these recent shifts are difficult to attribute to conventional causes. Instead, the recent shifts have been attributed to the introduction and adoption of Large Language Models (LLMs) \cite{Koppenburg2024,shapira2024delving,gray2024chatgpt,kobak2024delving,liang2024monitoring,liu2024towards,matsui2024delving,juzek2025does}. Examples of such spikes in word frequency concern words such as ``delve,'' ``intricate,'' and ``underscore.'' 

\begin{figure}[t]
    \centering
    \includegraphics[width=1\columnwidth]{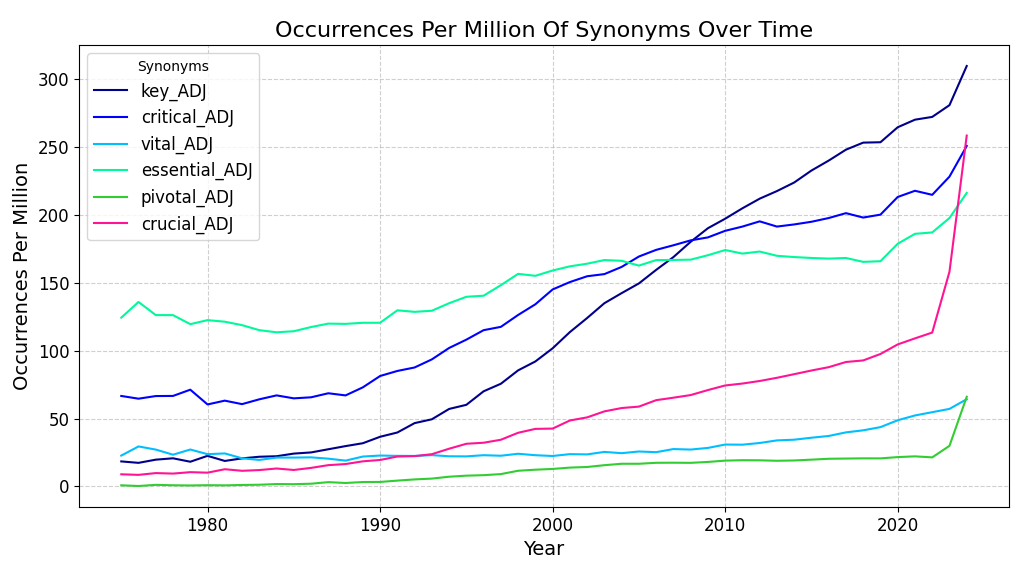}
    \caption{Trends in the occurrences per million words of the adjective ``crucial'' and five of its synonyms in PubMed abstracts over time. The large spike in frequency of ``crucial," identified in the literature as presumably due to AI, is followed, albeit less sharply, by its synonyms.}
    \label{fig:focalwordsopms}
\end{figure}

\begin{figure*}[t]
    \centering
    \includegraphics[width=1.8\columnwidth]{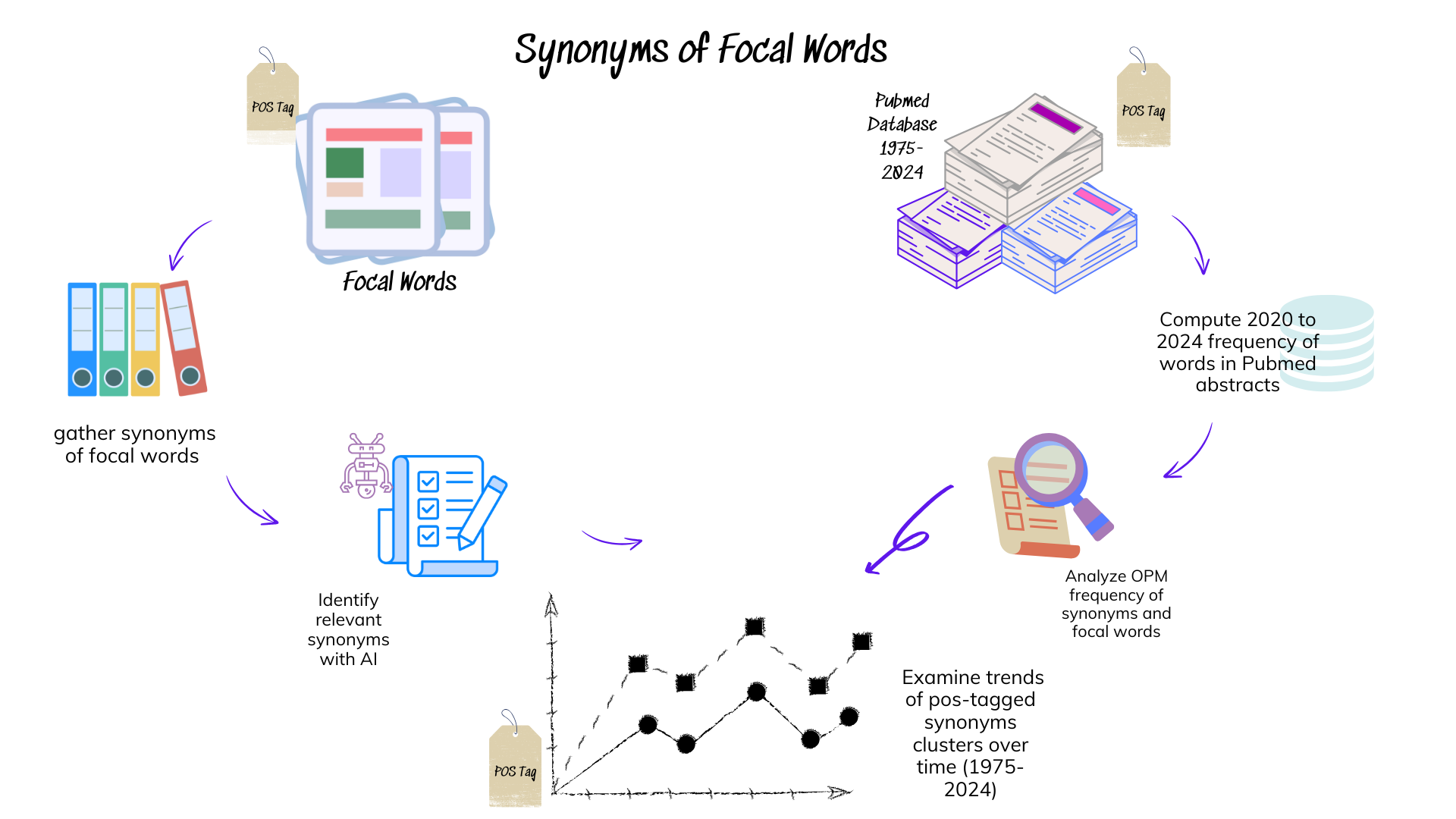}
    \caption{An illustration of the procedure used to systematically identify the synonyms of the focal words that are discussed in the literature as being overused by AI; figure created with Canva.}
    \label{fig:procedure}
\end{figure*}

While research has aimed to explain the reasons behind LLMs' overuse of certain words \cite{juzek2025does}, the exact nature of these spikes remains under-researched. Concretely, we identify two gaps in the current discourse. First, how do spiking words affect other lexical entries? Do they merely replace other lexical entries, or do they interact in more complex ways? Second, the existing literature has predominantly focused on positive spikes, that is, words that have sharply increased in relative frequency. Although explorative, this study systematically examines lexical entries that have declined in usage since LLMs became mainstream and explores whether these declines mirror the patterns observed in positively spiking words. Furthermore, previous studies have concentrated on full, inflected word forms, which collapse entries across syntactic categories (e.g. ``underscore'' could be a verb or a noun). The present work uses part-of-speech-tagged text as the basis for its analyses, which offers more adequate insights. 

The first aspect, how spiking words affect other lexical entries, is addressed in Section 2. Given that relative frequencies operate as a zero-sum dynamic, an increase in one lexical entry is necessarily linked to decreases elsewhere. In this context, we propose what we call `the replacement hypothesis' as the null hypothesis, where the rapid increase of one word is accompanied by a corresponding decrease in its semantic neighbors. However, our findings show that, in most cases, entire synonym groups tend to move collectively, invalidating the replacement hypothesis.

The second question, a systematic analysis of declining lexical items, arises from the observation that existing literature has predominantly focused on positively spiking entries. It is plausible to hypothesize that, just as LLMs strongly favor certain words, they may equally disfavor others. To address this, we apply a formal procedure to systematically investigate declining lexical items, as per Section 3. Our analysis of declining lexical entries reveals a more complex and less disruptive pattern.

We discuss the broader implications of these findings in Section 4. As Scientific English undergoes unprecedented linguistic change, understanding the structure and dynamics of these shifts is crucial for informing ongoing debates about the implications of recent developments and for evaluating the biases embedded in LLMs.

\section{2 Lexical Change in Focal Synonyms}
\label{sec:synonyms}

To enhance our understanding of lexical frequency changes driven by AI, we begin with the observation that the spikes reported in the literature are measured in relative terms, functioning as a zero-sum dynamic:\ An increase in one area necessitates decreases elsewhere. In this section, we turn our focus to the `elsewhere.' A plausible hypothesis is that a spiking lexical entry leads to decreases in words with similar meanings; for example, the rise of ``delve'' might coincide with declines in ``investigate,'' ``examine,'' and similar terms. We refer to this hypothesis as the replacement hypothesis, which serves as a kind of null hypothesis. In the following analysis, we critically examine the replacement hypothesis.

\subsection{Procedure}

To gain insights into the replacement hypothesis, we develop a rigorous procedure to identify synonyms of words known to be overused by LLMs, ``focal synonyms,'' parallel to the terminology used in Juzek and Ward (2025), who call spiking words that are AI-induced ``focal words.'' Starting from a list of 32 focal words widely discussed in the literature (as per Table~\ref{tab:synonyms1}; the original list and synonyms with frequencies can be found in our \href{https://github.com/fsu-nlp/structure-ai-lang-change}{GitHub repository}.
{https://github.com/fsu-nlp/structure-ai-lang-change}) due to rapid unexplained increases in usage, we convert each lexical item into its lemma form and its corresponding tag for part of speech (POS). We do so because certain entries in the literature are grammatically ambiguous. For example, when the literature discusses ``emphasizing,'' this could either be the adjective form (``emphasizing\_ADJ'') or the verb form (``emphasize\_VERB''). 

For each focal word, we use Merriam-Webster's online thesaurus to pull all of the synonyms available for the focal word under its corresponding POS \cite{merriamwebster}. For words and POS pairs without corresponding entries in Merriam-Webster, we include only the focal word and POS on our list. For example, ``comprehending\_ADJ'' is only listed in its verb form in the thesaurus. The list of each focal word, POS, and corresponding focal synonyms can be found in Table~\ref{tab:synonyms1} and in our \href{https://github.com/anon-authors3/Exploring-the-Structure-of-AI-Induced-Language-Change-in-Scientific-English/tree/main}{GitHub repository}. Many of these synonyms on Merriam-Webster, however, are not relevant in the context of academic literature. For example, synonyms of  ``delve\_VERB'' include ``explore,'' ``surf,'' and ``sift,'' only one of which is suitable. To remedy this with as little bias as possible, we use the OpenAI API in Python and query ChatGPT 4o-mini to evaluate the five most pertinent synonyms for each focal word (Prompt:\ ``I am looking for synonyms in the context of academic writing. Given the following {target\_pos} ``{target\_word}'', I have found these synonyms:\ {``, ''.join(synonyms)}. Which of these are most relevant in the context of the given part of speech in an academic abstract? Please disregard any repeats and only reply with a list of the 5 most relevant synonyms separated by commas (csv-like).''). 

\begin{figure}[t]
    \centering
    \includegraphics[width=1\columnwidth]{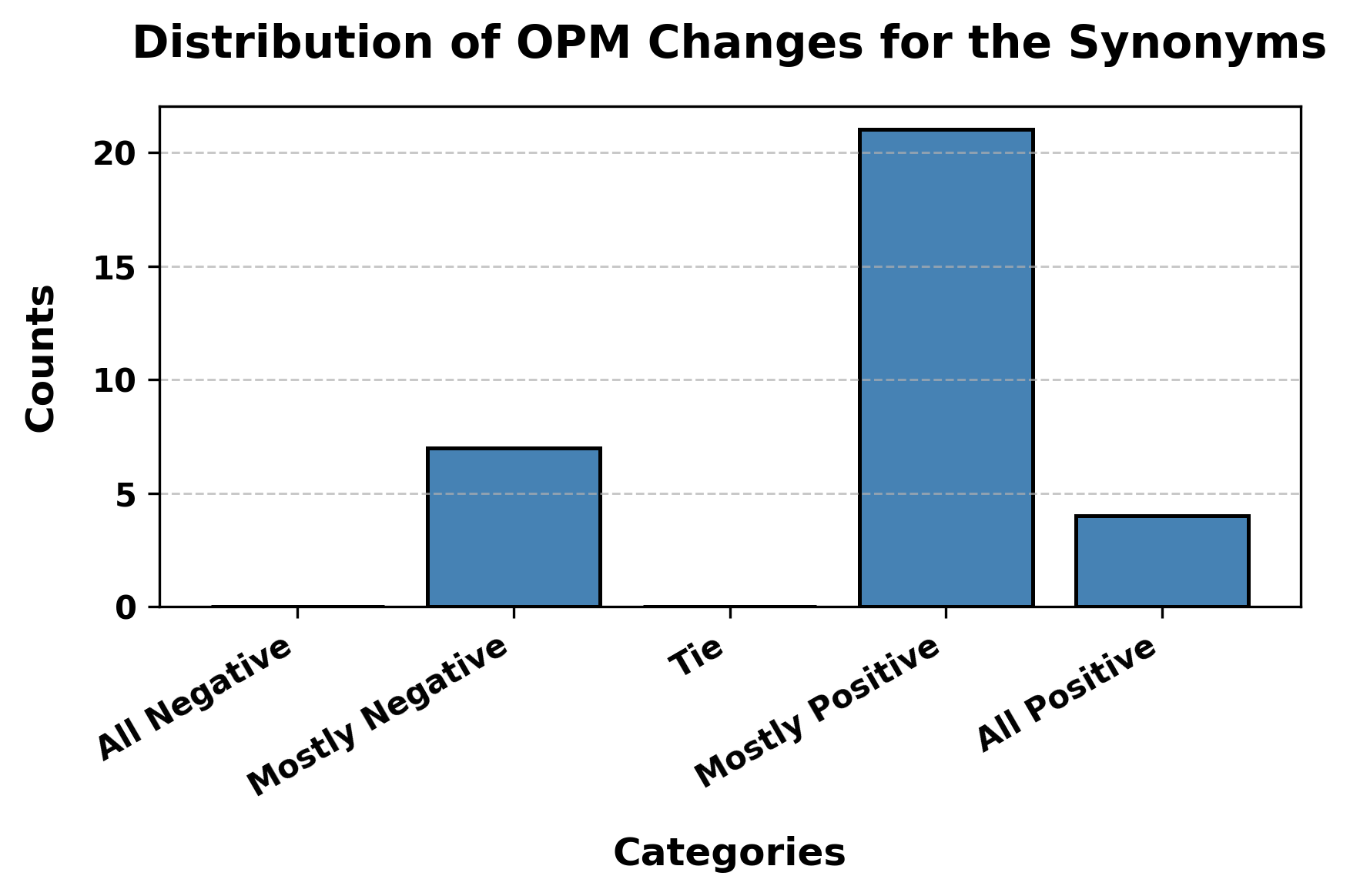}
    \caption{The distribution of how, for any given focal word, its four to five synonyms change in frequency between 2020 and 2024 (also see Table~\ref{tab:synonyms1} and our \href{https://github.com/anon-authors3/Exploring-the-Structure-of-AI-Induced-Language-Change-in-Scientific-English/tree/main}{GitHub repository}).}
    \label{fig:distribution}
\end{figure}

Taking an explorative approach to analyzing the frequency trends of the focal synonyms, we begin by following methods outlined in previous research for identifying word frequency trends in the literature. We used the PubMed public repository of scientific abstracts focusing on biomedical literature as our dataset \cite{PubMed}, downloaded through the PubMed API using a Python script. In addition, we used spaCy (model:\ encore\_web\_sm, version 3.8.0, 97\% accuracy, tagging of all data took about 140 hrs, done with a 2024 Thelio Custom machine with a Core i7-14700K), a POS-tagging library to identify the part of speech of each word in the abstracts analyzed \cite{montani2023spacy}. For comparison, we calculated the percent change of the POS-tagged words in occurrences per million (OPM; which is operationally equivalent to relative frequencies) from 2020 to 2024, and computed a chi-square contingency test for significance. The test is suitable for our data structure \cite{haslwanter2016introduction}, and we used the SciPy implementation to run it \cite{2020SciPy-NMeth}. This resulted in a list of words sorted from highest increase in frequency to highest decrease in frequency. We discarded words with POS tags representing punctuation, symbol, number, proper noun, and the ``other'' category (``X'') for clarity. In order to view long-term trends, and identify the magnitude of recent spikes or declines we conducted a corpus analysis, analyzing 28.8m Pubmed abstracts from 1975 to 2024, measuring the OPM for 6.94bn tokens. The entire procedure is illustrated in Figure~\ref{fig:procedure}.

\subsection{Analysis and Results}

To analyze the percentage change trends of only the focal synonyms, we compiled each focal word, followed by its synonyms and the resulting percentage change (see Table~\ref{tab:synonyms1} and in our \href{https://github.com/anon-authors3/Exploring-the-Structure-of-AI-Induced-Language-Change-in-Scientific-English/tree/main}{GitHub}). We noticed that in most cases, the synonyms of focal words also experience increases in frequency, as illustrated in Figure~\ref{fig:distribution}. Using the previous corpus analysis, we graphed the long-term trends of these synonyms in clusters of the focal word along with its corresponding synonyms, visually confirming these trends. An example graph can be found in Figure~\ref{fig:focalwordsopms}, which illustrates this for the word ``crucial'' and its synonyms, the remaining graphs can be found in our anonymous \href{https://github.com/anon-authors3/Exploring-the-Structure-of-AI-Induced-Language-Change-in-Scientific-English/tree/main}{GitHub repository}. 

Almost all of our focal synonyms, including synonyms of ``boast\_VERB,'' ``crucial\_ADJ,'' and ``pivotal\_ADJ,'' had multiple synonyms experiencing similar changes in frequency to the focal word. These trends may point to larger systematic changes happening within scientific English. It is known that LLMs show preferences to ``style words,'' that is, words that add little content in academic contexts \cite{kobak2024delving}. We note, however, that we do not think the low information density is inherent to the words themselves:\ ``crucial,'' ``meticulously,'' ``advancement'' could, in principle, be of high information content, if their meaning was supported by specific, real-world circumstances.

\subsection{The Value of Part of Speech Tags}

Due to differentiation of POS, we were able to distinguish between contexts of lemmas, giving further insight into how LLMs may be affecting language. Many of our focal words were verbs, adverbs, and adjectives, suggesting that the changes in Scientific English due to LLMs have little to do with base content (mainly nouns) and are clustered around extraneous language or AI-overused, low content words. 

\begin{figure*}[t]
    \centering
    \includegraphics[width=2\columnwidth]{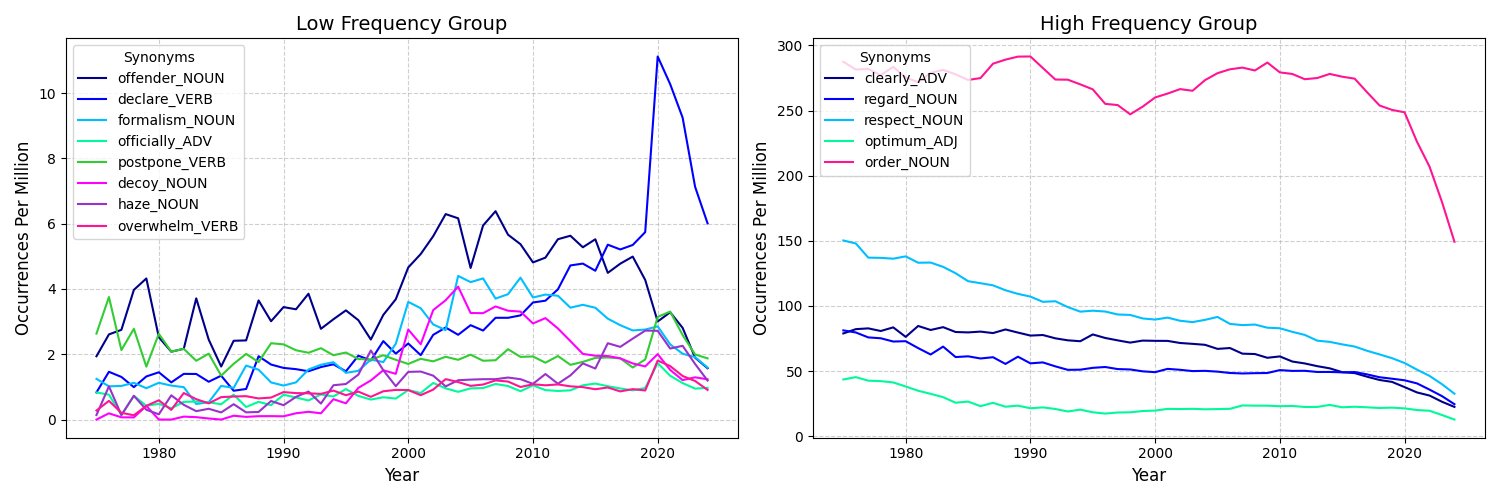}
    \caption{Trends in occurrences per million words for the 13 focal words in PubMed abstracts that displayed the largest decrease in usage between 2020 and 2024.}
    \label{fig:decreases}
\end{figure*}

Critically, we observe that, apart from ``advancement,'' nouns do not play any notable role. For the nouns we identified through the disambiguation of entries discussed in the literature, viz.\ ``underscore\_NOUN,'' ``boast\_NOUN,'' ``intricacy\_NOUN,'' and ``showcase\_NOUN,'' while they all show increases in usage, their initial frequencies are so low, well below 1 occurrence per million words, that they do not meaningfully contribute to the observed patterns. Similar observations apply to further disambiguations. For example, ``emphasize\_VERB'' displayed a large spike in frequency while ``emphasizing\_ADJ'' has a very low initial frequency (0.01 OPMs) and experienced very little changes in usage. By analyzing each word form separately, we can better understand what context the focal word is being used in.

The notable exception is ``potential\_NOUN,'' which increased from 333 OPMs in 2020 to 459 OPMs in 2024, with the adjective ``potential\_ADJ'' showing a similar rise from 476 OPMs in 2020 to 609 OPMs in 2024. In cases like ``potential,'' where multiple lexical entries across different POS categories are so important, the ability to distinguish between these entries is critical. These considerations illustrate the importance of POS tagging for such analyses.

\section{3 Corpus Analysis of Decreases}
\label{sec:corpusanalysis}

While most cases of our focal words exhibited synonym clusters displaying dramatic increases in frequency, we observed some focal synonyms in fact experienced a decrease in lexical frequency. One of such examples is the adjective ``important,'' a synonym for three of our focal words, ``significant\_ADJ,'' ``crucial\_ADJ,'' and ``noteworthy\_ADJ.''  ``important\_ADJ'' is notable as it displayed a significant decrease in frequency possibly due to AI. This implored us to raise the question:\ What is the structure of the decreases?

\subsection{Methodology}

To explore this, we built upon methods created to identify positively spiking items \cite{juzek2025does}, and adjust the methods such that we can identify decreases. For this, we examined the percent change and significance of all words pulled from the PubMed database. With the goal of identifying words whose sudden decrease in frequency was unexplained, two annotators independently analyzed this list from the bottom up leaving out any words that had obvious reasons for a rapid decline in usage (including words related to the Covid-19 pandemic which is now less prevalent, year numbers, or tokens whose POS tag did not align with the word). The annotators stopped when they reached 60 words, and in cases of disagreement the token was discussed (for the full list, see relevant files in our anonymous \href{https://github.com/anon-authors3/Exploring-the-Structure-of-AI-Induced-Language-Change-in-Scientific-English/tree/main}{GitHub repository}). In total, this resulted with a list of 72 words of lexical entries that potentially see a negative spike due to AI usage.

We say \textit{potentially}, because a negative spike in frequency itself is not enough to conclude LLM influence. For example, uses of the word ``amendment\_NOUN'' saw an almost 64 percent decrease in frequency from 2020 to 2024, however we found that ``amendment'' saw an increase in usage by AI. To identify words that were underused by both AI and human authors, we generated a random sample of 10 000 abstracts from the papers published in 2020 in the PubMed database and prompted ChatGPT 4.0 mini to summarize using key-words (Prompt:\ ``The following text is an abstract from a scientific paper:\ {input\_text} Summarize the abstract in keywords, separate keywords by commas''). The AI generated notes were then used to prompt ChatGPT to rewrite the abstracts (Prompt:\ ``Please write an abstract for a scientific paper, about 200 words in length, based on the following notes.'') In total, the AI generated 8187 abstracts (possibly due to topic sensitivity, ChatGPT would not provide a response for certain abstracts).

With the AI-authored abstracts we calculated the same OPM, change in frequency, and significance test as for the PubMed abstracts. We then examined if the 72 words on our list of unexplained decreases also saw significant decreases in the AI authored abstracts (see our our anonymous \href{https://github.com/anon-authors3/Exploring-the-Structure-of-AI-Induced-Language-Change-in-Scientific-English/tree/main}{GitHub repository}). If a word found on our original list experienced a decrease in the AI authored abstracts, it became a focal word. In total we recognized 13 decreasing focal words, including ``important\_ADJ,'' ``declare\_VERB,'' and ``clearly\_ADV'' (see Figure~\ref{fig:decreases} for the full list). 

\begin{table*}[t]
\centering
\begin{tabular}{|l|c|c|c|c|c|}
\hline
\textbf{Focal word} & \textbf{Synonym 1} & \textbf{Synonym 2} & \textbf{Synonym 3} & \textbf{Synonym 4} & \textbf{Synonym 5} \\
\hline
  meticulously\_ADV &     carefully &    rigorously &     precisely &      thoroughly & comprehensively \\
            +991.83\% (0.5 OPMs) &             -9.83\% &              +43.6\% &              +13.5\% &               +62.82\% &               +118.9\% \\
\hline
     showcase\_VERB &  demonstrate &   illustrate &       reveal &        exhibit &        present \\
            +508.12\% (3.2 OPMs) &             +20.22\% &            -13.88\% &             +28.92\% &               +75.68\% &               -6.92\% \\
\hline
   significant\_ADJ &   substantial &     important &       notable &      noteworthy &                    \\
             +27.42\% (934 OPMs) &             +53.18\% &            -23.43\% &            +263.69\% &              +153.51\% &                    \\
\hline
      surpass\_VERB &       exceed &    transcend &     outstrip &     outperform &        eclipse \\
            +347.91\% (4.1 OPMs) &             +33.78\% &             +94.52\% &            -29.34\% &              +103.21\% &                +8.48\% \\
\hline
   underscore\_VERB &    emphasize &    reinforce &    highlight &         stress &     accentuate \\
            +765.78\% (12.9 OPMs) &            +180.97\% &              +2.96\% &             +79.33\% &                -3.3\% &               +47.29\% \\
\hline
   underscore\_NOUN &     emphasis &    highlight &       stress &     importance &   significance \\
            +478.62\% (0.08 OPMs) &              +0.46\% &             +46.12\% &              +6.23\% &               +29.17\% &               +36.65\% \\
\hline
\end{tabular}
\caption{Changes in Occurrences per Million between 2020 and 2024 for a selection of focal words and their respective synonyms. For the focal words, the 2020 OPMs are given in parentheses. Remaining OPMs for 2020 vs.\ 2024 Occurrences per Million are omitted due to space limitations and can be found in our GitHub repo. The remaining focal words can be found in the \href{https://github.com/anon-authors3/Exploring-the-Structure-of-AI-Induced-Language-Change-in-Scientific-English/tree/main}{GitHub repository}.}
\label{tab:synonyms1}
\end{table*}

\subsection{Analysis and Results}

To examine the long-term trends of the decreasing focal words, we used the corpus analysis from 1975 to 2024 and analyzed the focal words changes over time. In contrast to what we had expected (a reverse of the drastic increasing spikes in word frequency) we observed more subtle, complicated trends (as per Figure~\ref{fig:decreases}). Many focal words had shown large fluctuations over years like ``fringe\_NOUN,'' and others displayed a sharp peak and then decline. This suggests that decreases in word frequency may not be so simple and as clear-cut as rapid increases, a point to which we will return in the discussion.

\section{4 Discussion and Conclusion}
\label{sec:conclusion}

Our findings expand upon the current understanding of language change due to LLMs. In Section 2, we explore the behavior and long-term trends of synonyms of words which see clustered spikes in frequency, giving insight into larger structural trends occurring. Clusters of synonyms experiencing rapid increases in frequency may point to syntactic change due to AI which may signify larger scale changes in scientific writing. Further, by analyzing the POS and synonyms of our focal words we were able to gain further insight into the structural changes of scientific language due to AI. All of our focal words were AI-overused, low content words. The rapid changes we have seen in Scientific English suggest that LLMs may be overusing these words, and in turn, changing the way scientific literature is written. From our findings, it is expected that abstracts that have been processed through LLMs contain higher frequencies of AI-overused, low content words than previous literature.

In the second part of our exploration, Section 3, we failed to find conclusive evidence of AI induced spikes in word frequency directly causing major decreases in frequency for other lexical items, finding instead a more complex relationship. This could be due to lexical usage tending to drop off at slower rates; novel innovations or social factors may cause an influx in new vocabulary that quickly increases in frequency, while terms generally fade to obsolescence \cite{aitchison2005language_change}. 

We do not observe widespread negative spikes that mirror the positive spikes reported in the literature (see references in the Introduction) for presumably AI-induced items; for an example of a positive spike, see Figure~\ref{fig:focalwordsopms}, which illustrates the increase in the usage of the word ``crucial.'' While it is important to exercise caution when drawing conclusions about the absence of a phenomenon, \textit{the absence of evidence is not evidence of absence} \cite{altman1995statistics}, our analysis systematically examines \textit{all} decreasing items. This allows us to confidently conclude that the pattern of decreases is broader and less systematic compared to the pattern of increases. Notably, many stylistic words exhibit spikes without any clear justification. Taken together, our results suggest that the collective increases of focal words and their synonyms, along with the absence of sharp decreases, indicate that words such as ``advancement,'' ``emphasize,'' and ``meticulous'' do not simply replace similar terms. Instead, their usage reflects semantic and pragmatic additions. ``crucial'' does not replace ``critical,'' or ``essential'', they all are added, arguably as a qualification. The increase ``underscore'' co-occurs with an increase of ``emphasize'' and ``highlight,'' adding a pragmatic choice to one's writing.

\subsection{Limitations}

Our research is partially limited in terms of variety, linguistic level, lexical items, and genre. While the scientific abstracts sampled from PubMed provide a concise summary of linguistic patterns, further lexical choices may be found in the body sections of the articles. Furthermore, while PubMed contains articles from many different scientific fields, the focus is on biomedical and life sciences. We focus on academic English and the lexical items the literature discusses in this context (as per references in the introduction), as Large Language Models exhibit domain-specific behavior; they are context-sensitive.  Future research should continue to probe the exact structure of the changes that are occurring, focusing on other varieties, even other languages, as well as other linguistic levels (e.g., syntax changes), and a broader range of scientific disciplines to assess the generalization of these findings.

It is also possible that the observed shifts in word frequency are driven by the topics published at a given time, such as ``SARS'' or ``Covid-19'' in recent years, which saw increases resembling step functions. However, no underlying world event, except the introduction of LLMs, correlates with the sudden linguistic shifts observed in 2023. An alternative explanation could be a widespread shift in writing style within abstracts not caused by LLM usage, but this is improbable given the timing and extent of the changes observed. 
 
Furthermore, the frequency shifts observed are often attributed, by conjecture, to the influence of LLM usage. While this is a plausible assumption grounded in existing literature, it remains an assumption. To definitively prove that these changes are directly caused by LLM usage, or by their broader influence, requires additional investigation.

It is also important to consider the various mechanisms through which LLMs may exert influence. One possibility is the direct use of LLM-generated language; another possibility is an indirect effect. LLM is ubiquitous and its patterns might have directly entered the language faculty of some of the PubMed authors. More research is necessary to gain further insights. For this, it would be valuable to analyze spoken language. This has been done for scripted spoken language \cite{geng2024impact,yakura2024empirical}, but we would also want to see this done for unscripted spontaneous speech. Further, A longitudinal analysis of spontaneous spoken language, compared with trends in written language, could offer insights into the question of causality:\ if certain changes in written language, presumably influenced by AI tools, consistently precede similar patterns in spoken language, then this would support the interpretation that AI-driven language is actively influencing human linguistic behavior.

A major question, beyond the scope of this paper, remains:\ \textit{Why} do language models exhibit the observed patterns of over- and underuse? It was noted that the overused words are not disproportionately frequent in the training data \cite{liang2024mapping,juzek2025does}. One plausible possibility is that the (Reinforcement) Learning from Human Feedback stage may be a key contributor to such lexical biases \cite{juzek2025does}. Supporting this view, it was found that preference-tuned LLMs are more inclined toward stylistic features such as boldface text and bullet lists compared to their base model counterparts \cite{zhang2024lists}.

\subsection{Broader Impacts}

Scientific English is undergoing unprecedented language change. Our observations indicate that these changes are not limited to individual lexical entries but extend to entire lexical groups. This pattern is consistent with the interpretation that, if these changes are indeed LLM-induced, they reflect pragmatic adjustments rather than simple lexical replacements. In our analyses, we considered part-of-speech categories for more insightful results.  Additionally, we systematically explored the structure of decreasing lexical entries.

Our work is important, as it helps us understand the magnitude and structure of technology's impact on human language, contributing to the broader discourse on whether these changes are desirable or undesirable. If deemed undesirable, the differences in LLM language production could be considered a form of bias or misalignment. Further, these discrepancies from past writing may lead to the homogenization of scientific literature by creating biases towards a specific style of writing. Our work certainly makes clear that the changes go deeper than just a few lexical items.

We observe that technology companies, such as OpenAI, have considerable influence on human language. With this influence comes a responsibility to consider the broader implications of their impact. A critical first step, to which our work contributes, is to systematically map the extent of this influence, providing the foundation for an informed discussion on the ethical and social responsibilities of these companies.

\subsection{Concluding Remarks}

Our research adds to the current understanding of how AI has begun to change Scientific English, and there is no doubt that the existence of LLMs will have many more large-scale impacts over time. The synonym clusters we observe point to changes much larger than single word bias as found in prior research, leading us to wonder how these short-term trends will propagate over time leading to less accurate models and perhaps much more language change \cite{alemohammad2023self,briesch2023large,hataya2023will}. 

Alternatively, these changes could be seen as a temporary `learning curve' and LLMs will adapt to better human sentence structure over time. As LLMs become more widespread it is essential to be aware of the broader impacts they may have on all areas of human existence, including language. 

\subsection{Ethical Considerations}

We used AI-assisted programming tools (GitHub Copilot and ChatGPT 4o) for the development of our code. All code was reviewed by the authors.
This research does not contribute to overgeneralization, confirmation bias, or the disproportionate representation of specific languages, topics, or applications at the expense of others.

\subsection{Acknowledgments}

The authors thank the reviewers for their thoughtful, constructive, and kind feedback. 

\bibliographystyle{flairs}
\bibliography{custom}

% \section{Appendix:\ \href{https://github.com/anon-authors3/Exploring-the-Structure-of-AI-Induced-Language-Change-in-Scientific-English/tree/main} All Target Words and Their Synonyms}

\end{document}